\documentclass[twoside]{article}
\usepackage{PRIMEarxiv}
\usepackage[utf8]{inputenc} 
\usepackage[T1]{fontenc}    
\usepackage{hyperref}       
\usepackage{url}            
\usepackage{booktabs}       
\usepackage{amsfonts}       
\usepackage{nicefrac}       
\usepackage{microtype}      
\usepackage{lipsum}
\usepackage{listings}
\usepackage{fancyhdr}       
\usepackage{graphicx}       
\graphicspath{{media/}}     
\usepackage{subcaption}
\usepackage{calc}
\usepackage{amssymb}
\usepackage{amstext}
\usepackage{amsmath}
\usepackage{amsthm}
\usepackage{multicol}
\usepackage{pslatex}
\usepackage{apalike}
\usepackage{algorithm2e}
\usepackage[bottom]{footmisc}
\usepackage{multirow}
\usepackage{xcolor}
\usepackage{verbatim}

\usepackage{color}
\usepackage[dvipsnames]{xcolor}
\usepackage{multirow}
\usepackage{amsmath} 
\usepackage{pgfplots}
\pgfplotsset{compat=1.18} 
\usepackage{tikz}

\begin{document}

\fancyhead[LO]{A symbolic algorithm for the unification of Nawatl word spellings}
\fancyhead[RE]{Guzman-Landa et al.} 

\title{A symbolic Perl algorithm for the unification of Nahuatl word spellings}

\author{Juan-José Guzm\'an-Landa$^1$, Juan-Manuel Torres-Moreno$^1$, Graham Ranger$^2$\\
$^1$LIA \& $^2$ICTT, Avignon Université\\Avignon, France\\
\texttt{\{juan-jose.guzman-landa, juan-manuel.torres, graham.ranger\}@univ-avignon.fr}
\AND
Ligia Quintana-Torres$^3$, Miguel Figueroa-Saavedra Ruiz$^4$, Martha-Lorena Avenda\~no-Garrido$^3$\\ 
$^3$Fac. de Matem\'aticas \& $^4$Inst. de Investigaciones en Educación,\\
Universidad Veracruzana, Xalapa, Mexico\\
\texttt{\{migfigueroa, ligiaquintana, maravendano\}@uv.mx}
\AND
Jesús Vázquez-Osorio, Gerardo Eugenio Sierra Mart\'inez\\
GIL, Universidad Nacional Aut\'onoma de México,\\
Coyoac\'an, CDMX, Mexico
\\ \texttt{gsierram@iingen.unam.mx}
\AND
Patricia Vel\'azquez-Morales\\
Avignon, France
}

\maketitle              
\begin{abstract}
In this paper, we describe a symbolic model for the automatic orthographic unification of Nahuatl text documents. 
Our model is based on algorithms that we have previously used to analyze sentences in Nawatl, and on the corpus called $\pi$-{\sc yalli}, consisting of texts in several Nawatl orthographies.
Our automatic unification algorithm implements linguistic rules in symbolic regular expressions. 
We also present a manual evaluation protocol that we have proposed and implemented to assess the quality of the unified sentences generated by our algorithm, by testing in a sentence semantic task. We have obtained encouraging results from the evaluators for most of the desired features of our artificially unified sentences.
\end{abstract}


\section{Introduction: the challenges for NLP of Nawatl}

Nawatl, also known as Nahuatl or Mexicano,\footnote{In this work, although there are other variations of this name are \textit{nauatl}, \textit{nawat}, and \textit{naualajtoli}, we use the term Nawatl because it is currently the most used term among Nahuan writers and fits in with the alphabetic reforms of our target population. Other names specific to the language—and according to its dialectal and diagraphic variation—are \textit{mexikano}, \textit{mejikano}, \textit{mexcatl}, \textit{mexkatl}, masewaltlahtolli, \textit{maseualta'tol}, \textit{masehualtla'to}l, \textit{maseualtajtol}, \textit{melatajtol}, and \textit{mela'tajtol} \cite{INALI}.} is a language with limited computational resources, despite being a living language spoken by approximately two million people.
The newly introduced $\pi$-\textsc{yalli} corpus provides a foundation for research and development in Machine Learning and Artificial Intelligence.
However, the presence of multiple orthographic systems across dialectal varieties and writing traditions of Nawatl \cite{saavedra2024amapowalistli} poses significant challenges for machine learning.
This issue underscores the inherent complexity of working with this $\pi$-language.
To address this challenge, we propose a novel symbolic unifier for Nawatl orthographies, grounded in well-established linguistic principles, to facilitate text processing.

Nawatl is an indigenous language belonging to the Uto-Nahuan language family \cite{smith2002aztecs,parlons}. It is spoken by a substantial number of individuals in Mexico and other regions of the Americas\footnote{\url{https://en.wikipedia.org/wiki/Nahuatl}}.
Nawatl has been spoken in Mesoamerica since the 5th century, and remains to this day the most widely spoken national language in Mexico after Spanish, with 1,651,958 speakers recorded in the most recent census \cite{inegi2020censo}, and a total of over 2.5 million individuals within the broader Nawatl-speaking community.
According to INALI,\footnote{\url{https://www.inali.gob.mx/clininali/html/l\_nahuatl.html}} there are 30 recognized linguistic varieties of Nawatl spoken across Mexico.
Currently, several of these varieties are classified as endangered \cite{unesco2012atlas}.
This situation persists despite continuous efforts made by the Nawa communities since 2003—this year, Nawatl was officially recognized as a national language—to promote its use in both oral and written forms across various domains such as publishing, higher education, mass media, and digital platforms \cite{aguilar2023tecnologias,saavedra2023nawatlahtolli}.
As a polysynthetic and agglutinative language, Nawatl exhibits a complex morphological structure wherein words are composed of nominal or verbal roots combined with various morphemes that contribute to the overall meaning. An example of agglutination is \textit{cuauhtochtontli} (Little wild rabbit) that is composed of \textit{cuahuitl} (tree/wood), \textit{tochtli} (rabbit), -\textit{ton} (diminutive), and \textit{-tli} (noun): cuauh+toch+ton+tli.

Syntactic relations between words are primarily established through verb valency and grammatical particles. These particles themselves may be morphologically complex, conveying subtle semantic distinctions in addition to fulfilling discursive functions.
Certain lexical forms function as ``sentence-words'', incorporating subject and predicate information along with grammatical markers for actants, modality, relationality, and directionality within a single morphological unit. An example of polysynthesis is the sentence "Axcan timitzoncochtiah" (Now we cause you to sleep far away). Here the word \textit{timitzoncochtiah} is itself a sentence "we cause you to sleep far away": ti- (we), mitz- (to you), on- (directional, far away), cochi (sleep), -tia (causative, cause to do the action), and -h (plural verb form).
Rather than describing Nawatl as a minority language—a term that may carry negative connotations—we prefer to adopt the term $\pi$-language, referring to languages with limited computational resources \cite{these-pi,abdillahi:hal-01311495}.\footnote{In contrast to $\tau$-languages, which are well-resourced, and $\mu$-languages, which are moderately resourced in computational terms.}
Despite its rich linguistic and cultural heritage, Nawatl faces significant challenges due to its status as a $\pi$-language and the scarcity of computational tools available for its preservation and revitalization.

\color{black}

However, the creation of new computational resources requires automatic natural language processes in Nawatl that do not exist. 
To our knowledge, there are no POS parsers or taggers, no stemmers or lemmatizers and even less a tokenizer.
The few corpora available are small, and the texts they comprise are in several linguistic variants, with little or no orthographic standardization. The aim of this paper is to build a Nawatl spelling unifier that will standardize the available textual resources.
We will show how the use of unigraphy enables us to increase the performance of the models in a classic and fundamental NLP task: the calculation of semantic similarity.

This article is organized as follows: in Section~\ref{sec:poly}, we introduce the issue of Nawatl polyorthography.
In Section~\ref{sec:modelo_linguistico}, we describe our linguistic model for orthographic unification.
Section~\ref{sec:experiments} presents the pilot Nawatl corpus and the evaluation protocol for recent algorithms (LLM vs. traditional LM approaches), and we report our experimental results, followed by conclusions and future directions in Section~\ref{sec:CONC}.




\section{Polyorthography: a challenge for the computational processing of corpora}
\label{sec:poly}

\begin{figure}
    \centering    \includegraphics[width=0.85\linewidth]{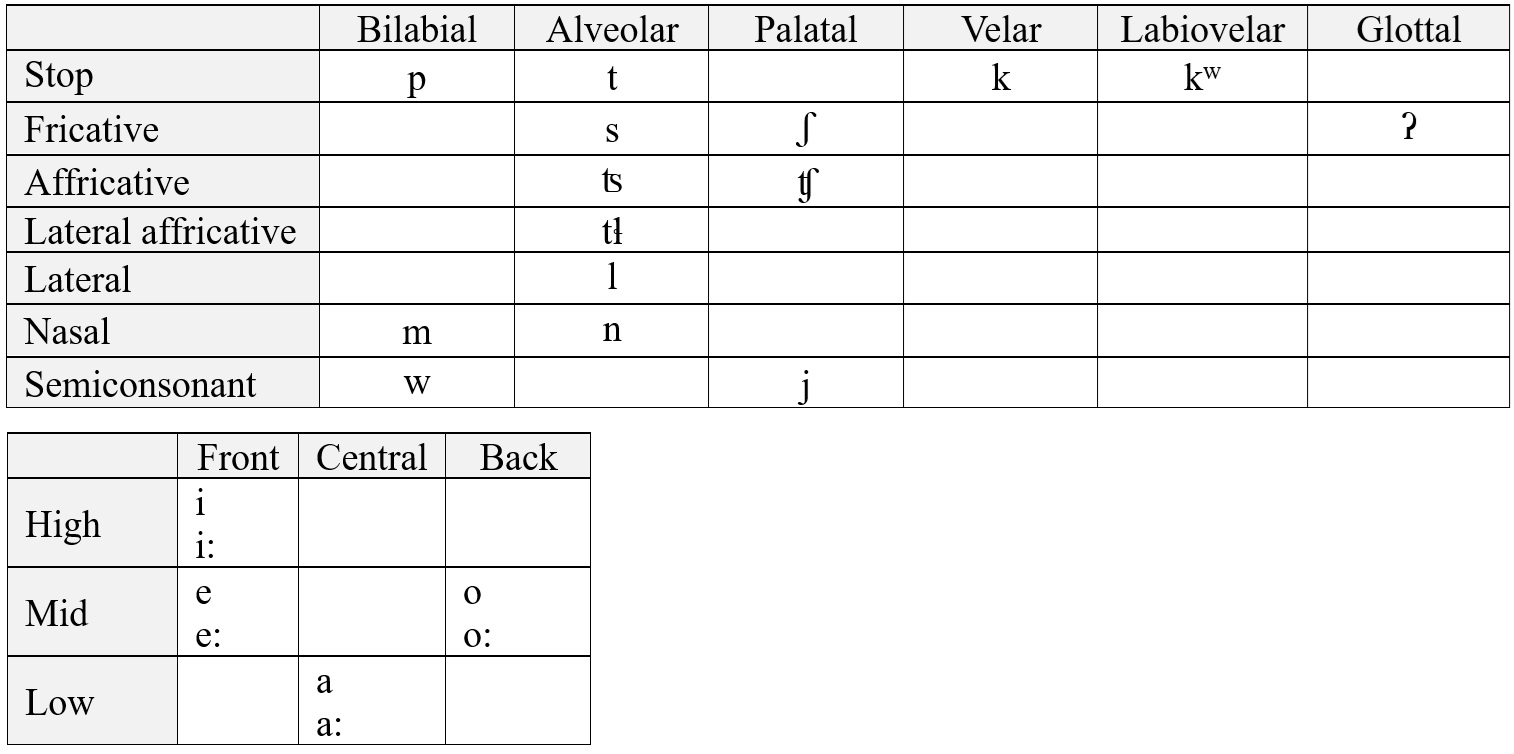}    
    \caption{Top: Consonant Phonemes in the Nawatl Language. Bottom: Vowel Phonemes in the Nawatl Language \cite{monzon}.}
    \label{fig:conson-vocal}
\end{figure}

With respect to the writing system of Nawatl, several clarifications are necessary. Since the 16th century—when the Nawatl-speaking community replaced its semasiographic writing system with an alphabetic system (initially influenced by Latin and Spanish, and later by the International Phonetic Alphabet)—various standardized alphabets have emerged.
Although Nawatl features 8 vowel phonemes and 15 consonant phonemes (see Fig. \ref{fig:conson-vocal}), each phoneme is not necessarily represented by a single letter (grapheme). In many cases, individual phonemes have been represented using different letters to reflect allophonic variation, or alternatively, by means of digraphs, diacritical marks, or even without a letter in some writing systems.

\begin{figure}[ht!]
    \centering    \includegraphics[width=0.8\linewidth]{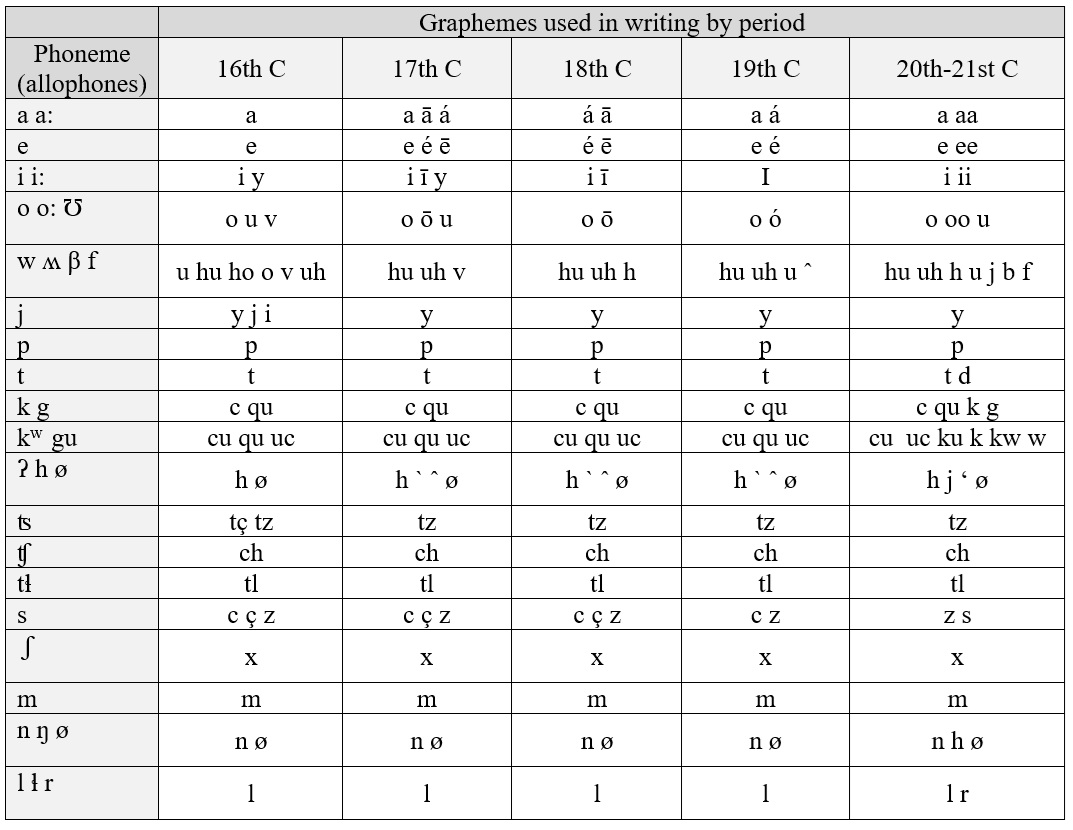}
    \caption{Phoneme–Grapheme Correspondences in Nawatl Writing. Tableau inspired from \cite{saavedra2024amapowalistli}.}
    \label{fig:phon-graph}
\end{figure}

Moreover, the same graphemes have not always been used consistently. Over time, they have undergone graphical changes, influenced by the evolution of Spanish orthography or through independent adaptations aimed at reducing the number of letters and at simplifying digraphs. Historically, this has led (see Fig.~\ref{fig:phon-graph}) to a wide diversity of possible representations for writing the language.
This variability is not only the result of differing phoneme representations, but also stems from the phonological, lexical, and morphosyntactic variation expressed across the dialectal varieties of Nawatl, which affects word formation and usage.

Although this polynormative and polyorthographic system does not pre\-sent difficulties for speakers and writers of the language in terms of information transmission, it does pose a significant challenge for machine learning. It highlights the need for tools capable of simplifying and unifying orthographic variation to enable effective automatic recognition and processing of Nawatl texts.
Therefore, in the following section, we propose a unified orthographic system designed to facilitate the computational processing of Nawatl.

\section{A Linguistic-Symbolic Poly-orthographic Uni\-fi\-ca\-tion Model}
\label{sec:modelo_linguistico}

Our proposal for the unification of polyorthographies is based on the analysis of letters and digrams, with the help of a Nahuatl specialist, being the fifth author of this article. All in order to simplify or merge particular cases. Thus, in Table \ref{tab:unigraph} we present the rules that standardize these particular cases. For example, in the case of the phoneme /w/ this has been represented with the letters hu, uh, u and w. In this case we have reduced it to the grapheme w (Table \ref{tab:unigraph}, rules 1 and 2): cases like \textit{huehuechiuhtiazquia} to \textit{wewechiwtiaskia}, \textit{huehuetque} to \textit{wewetke}, \textit{huehuey} to \textit{wewey}, among others. Another example would be the phoneme /s/ which was represented with the letters c, ç, z and s before the vowels i and e, or at the end of the word with z or s, in which case we reduce it to s (Table \ref{tab:unigraph}, rules 15, 21 and 22): \textit{çioatl} to \textit{sioatl}, \textit{çiva} to \textit{siva}, \textit{çihuateupixque} to \textit{siwateupixke}.

We also process certain exceptions (some of them paleographic) in order to reduce the variability of the texts.

Table~\ref{tab:unigraph} presents our proposed set of unigraphic rules, based on the linguistic considerations and observations outlined in Section~\ref{sec:poly}.
The pseudo-linguistic patterns are written in a way that facilitates direct transcription into regular expressions.
%

\begin{table}[ht!]
\centering
 \begin{tabular}{|c|c|c|}
\hline
 \bf Rule & \bf Pattern  & \bf Replacing \\ \hline
 1 & hu+vowel & hu$\rightarrow$w\\ 
\hline
 2 & vowel+uh & uh$\rightarrow$w\\ 
\hline
 3 & qua & q$\rightarrow$k\\
\hline
 4 & que|qui & qu$\rightarrow$k\\
\hline
 5 & (vowel|\texttt{<sp>})+u+vowel & u$\rightarrow$w\\ 
\hline
 6&ca|co|cu &c$\rightarrow$k\\ 
\hline
 7&kw &w$\rightarrow$u\\ 
\hline
 8 &(uk$|$ku)+vowel & c$\rightarrow$k\\ 
\hline
 9 & c $\notin$ ch & c$\rightarrow$k\\ 
\hline
 10 & consonant+y+letter & y$\rightarrow$i\\ 
\hline
 11 & vowel+j+(vowel|consonant|\texttt{<sp>}) & j$\rightarrow$h\\ 
\hline
 12&ywan&ywan$\rightarrow$iwan\\
\hline
 13 & \texttt{<sp>}+wan+\texttt{<sp>} & wan$\rightarrow$iwan\\
\hline
 14 & yn & yn$\rightarrow$in\\
\hline
 15 & ce | ci | çi | zi & [ c ç z ]$\rightarrow$s\\
\hline
 16 & zo & z$\rightarrow$s\\
\hline
 17 & tz	| tc | tç & [ c ç z ] $\rightarrow$ s\\
\hline

17' & zt | ct | çt  & [ c ç z ] $\rightarrow$ s\\
\hline
 18 & {\bf \color{black}{ll}	| hl} & ( ll $|$ hl ) $\rightarrow$ l\\
\hline
 19 & \texttt{<sp>}+{\bf \color{black} i}+vowel & i$\rightarrow$y\\
\hline
 20 & (\texttt{<sp>}$|$consonant)+{\bf  u}+consonant & u$\rightarrow$o\\
\hline
 21 & za	| ça & [ ç z ] $\rightarrow$ s\\
\hline
 22 & ze	| çe & [ ç z ] $\rightarrow$ s\\
\hline
 23 & \texttt{<sp>}+(ypan|pan)+\texttt{<sp>}& ypan $\rightarrow$ ipan\\
\hline
 24 & \texttt{<sp>}+(wan|tech)+\texttt{<sp>}& wan|tech $\rightarrow$ iwan | itech\\
\hline
 25 & (aa|ee|ii|oo) & $\rightarrow$ a | e | i | o\\
\hline
 26 & (ha|he|hi|ho) & $\rightarrow$ ah | eh | ih | oh\\
\hline
 27 & \texttt{<sp>}+(ahmo|hamo|ajmo)+\texttt{<sp>}& $\rightarrow$ amo\\
\hline
 28 & vowel+c+hu+vowel & chu $\rightarrow$ kw\\
\hline
 29,31,32 &(ã|à|á|ā)|(è|é|ē|\~e)|(í|ī|ì)|(ò|ó|ō) & $\rightarrow$ a | e | i | o\\
\hline
 30 &â|ê|î|ô & $\rightarrow$ ah | eh | ih | oh\\
\hline
 33 & (Ç ç) | (\~y ŷ ÿ) & $\rightarrow$ s | y\\ \hline
 34 & ihua[n] & ihua[n]→ihuan paleographie \\ \hline
 35 & vowel:word & a:tl→atl paleographie\\
\hline

\end{tabular}
\caption{Linguistic Rules of the Proposed Unigraphic System. ``|'' = or, \texttt{<sp> = !?;.}}
\label{tab:unigraph}
\end{table}

\section{Experiments: two semantic similarity tasks}
\label{sec:experiments}

We introduced two pilot datasets to establish an evaluation protocol.
To this end, two semantic similarity tasks were explored: semantic similarity between words and semantic similarity between sentences.
For our similarity experiments, we used the openly available Nawatl corpus $\pi$-\textsc{yalli}.\footnote{Nawatl $\pi$-\textsc{yalli} corpus can be accessed at: \url{https://demo-lia.univ-avignon.fr/pi-yalli}}
This corpus was designed to train machine learning models (static or dynamic, but not generative), enabling the development of computational resources for the Nawatl language with more than 6 million tokens \cite{NAHU2}.

\subsection{Task I: Semantic Similarity Between Words}

The first experiment focuses on semantic similarity between words (and sentence-words) in Nawatl.
This is a classic task in Natural Language Processing (NLP).
Given 23 reference terms, each associated with a list of 5 candidate terms, 27 native Nawatl-speaking annotators were asked to semantically rank the candidates, from the closest to the most distant in meaning relative to the reference \cite{NAHU2}.\footnote{An odd number of annotators was chosen to prevent potential tie rankings.}
Each candidate was assigned a semantic score from 1 to 5 (1 being the most similar to the reference, and 5 the least). This resulted in a total of 23$\times$27 = 621 rankings, each containing 5 words. The rankings were then aggregated into a consensus ranking.
This pilot dataset, annotated by Nawatl speakers from various linguistic regions, was used to assess the diversity of possible orthographies and the challenges of simultaneously processing multiple dialectal varieties of Nawatl.
It also served to evaluate the quality of language models (ML) trained on the $\pi$-\textsc{yalli} corpus. 
A detailed description of this task is available in \cite{NAHU2}.
Table~\ref{tab:tau_test_LLM} shows the performance of several LLMs and ours best statics LMs on the word-level semantic similarity task (from \cite{piyalliTALN}). 
The best-performing model was Gemini 2.0, with a Kendall’s $\tau$ score of 0.696.

\begin{table}[ht!]
  \centering
  \begin{tabular}{|c|c|}
\hline
  {\bf Pretrained LLM and trained LM} & \bf ~Kendall's $\tau$~~\\ \hline
  Gemini 2.0 &\bf 0.696  \\ \hline  
  Claude 3.7 & 0.644 \\ \hline
  DeepSeek V3 &  0.522 \\ \hline
\textbf{\textit{Glove 300D enhaced unigraphy}} & \textbf{\textit{0.496}} \\  \hline
  ChatGPT-4 mini & 0.487 \\  \hline 
\textbf{\textit{FastText 300D unigraphy}} & \textbf{\textit{0.469}} \\  \hline 
  Grok 3& 0.452  \\ \hline
  Copilot & 0.399  \\ \hline
\textbf{\textit{FastText 300D raw text}} & \textbf{\textit{0.362}} \\  \hline 
  Mistral Large& 0.226 \\ \hline
  Llama-3.3-70B-Instruct& 0.107  \\ \hline
  \end{tabular}  
   \caption{Word Semantic Similarity, Kendall’s $\tau$: LLMs vs. Consensus Ranking.} 
 \label{tab:tau_test_LLM}
\end{table}

Table~\ref{tab:tau_test_sansUNI} presents the details of results of the static models Word2Vec \cite{Mikolov2013distributed}, FastText~\cite{bojanowski-etal-2017-enriching}, and GloVe~\cite{glove}, trained with varying dimensions ranging from 50 to 300, using the non-unified (non-unigraphic) corpus.
These results are those reported by~\cite{piyalliTALN}.
The best-performing static model was FastText CBOW, which achieved a Kendall’s $\tau$ score of 0.362.

\begin{table}[ht!]
\centering
\begin{tabular}{|c|c|c|c|c|c|}
  \multicolumn{6}{c}{\bf Trained Local Embeddings, epochs=20, window=5} \\ \hline
  \multirow{3}{*}{\bf $D$}& \multicolumn{2}{c|}{\bf Word2Vec} & \multicolumn{2}{c|}{\bf FastText} & \multirow{2}{*}{\bf Glove}\\ 
  \cline{2-5}
  &CBOW & Skip-Gram & CBOW  & Skip-Gram & \\ 
  \cline{2-6}
  & \multicolumn{5}{c|}{$\tau$; $\pi$-{\sc yalli} {\bf raw corpus}}\\
  \hline
  50   & 0.223 & 0.293 & 0.322 & 0.299 & 0.197 \\ \hline  
  100  & 0.191 & 0.249 & 0.351 & 0.362 & 0.171 \\ \hline
  150  & 0.171 & 0.206 & 0.336 & 0.304 & 0.197 \\ \hline  
  200  & 0.258 & 0.223 & 0.354 & 0.296 & 0.197 \\ \hline  
  300  & 0.180 & 0.232 &\bf 0.362 & 0.296 & 0.171 \\ \hline 
 \end{tabular}
   \caption{Word Semantic Similarity, Kendall’s $\tau$: Static Embeddings vs. Consensus Ranking on the Raw Corpus
(Values reported in \cite{piyalliTALN}).} 
 \label{tab:tau_test_sansUNI}
\end{table}

Table~\ref{tab:tau_test_UNI} shows the details of the results of same static models—Word2Vec, FastText, and GloVe— this time applied to the $\pi$-\textsc{yalli} corpus and the pilot word semantic similarity dataset, both preprocessed using the proposed unigraphic normalization \cite{piyalliTALN}.
The best-performing static model is once again FastText with the CBOW architecture, now achieving a Kendall’s $\tau$ score of 0.469.
The preprocessing step involving unigraphic normalization yields a performance improvement of over 29.5\% compared to the previous results on the raw corpus.

\begin{table}
\centering
\begin{tabular}{|c|c|c|c|c|c|}
  \multicolumn{6}{c}{\bf Trained Local Embeddings, epochs=20, window=5} \\ \hline
  \multirow{3}{*}{$D$} & \multicolumn{2}{c|}{\bf Word2Vec} & \multicolumn{2}{c|}{\bf FastText} & \multirow{2}{*}{\bf Glove}\\ 
  \cline{2-5}
  &CBOW & Skip-Gram & CBOW  & Skip-Gram & \\ 
  \cline{2-6}
   &\multicolumn{5}{c|}{$\tau$ ; $\pi$-{\sc yalli} \textbf{unigraphied}} \\
  \hline
  50  & 0.220 & 0.212 & 0.368 & 0.328 & 0.273 \\ \hline  
  100  & 0.246 & 0.278 & 0.399 & 0.394 & 0.325 \\ \hline
  150  & 0.244 & 0.252 & 0.429 & 0.336 & 0.336 \\ \hline  
  200  & 0.287 & 0.339 & 0.429 & 0.362 & 0.359 \\ \hline  

  300  & 0.269  & 0.278 & \bf 0.469 & 0.365  & 0.339  \\ \hline
 \end{tabular}
 \caption{Word Semantic Similarity, Kendall’s $\tau$: Static Embeddings vs. Consensus Ranking with Unigraphic Normalization
(Values reported in \cite{piyalliTALN}).} 
 \label{tab:tau_test_UNI}
\end{table}

\subsubsection{Enhanced Unigraphic}

In light of the previous results, we decided to implement an improved preprocessing pipeline for the $\pi$-\textsc{yalli} corpus. In particular, the enhanced preprocessing included:

\begin{enumerate}
\item enriched unigraphic normalization, covering all accented vowels and consonants (both uppercase and lowercase);
\item the removal of  paleographic symbols found in 16th-century texts;
\item the elimination of duplicate phrases in the corpus;
\item the normalization of specific lexical exceptions not resolved by the unigraphic rules.\footnote{For example: {miak, miyak} $\Rightarrow$ {\bf miyak} (‘many’); {huei, huey, uei} $\Rightarrow$ {\bf wey} (‘great/large’).}
\item the elimination of two more frequently stopwords: \textit{iwan} and \textit{in}. 
\end{enumerate}

Table~\ref{tab:tau_test_UNI+} reports the results of the static models Word2Vec, FastText, and GloVe, restricted to 300 dimensions (as this setting yielded the best results in Table~\ref{tab:tau_test_UNI}), using the improved preprocessing and unigraphic normalization on the $\pi$-\textsc{yalli} corpus and the pilot dataset for the task.

The models FastText CBOW/Skip-gram and Glove on Table \ref{tab:tau_test_UNI+} outperform the results with respect to Table \ref{tab:tau_test_UNI}.
Once again, the best-performing static model is Glove, now achieving a Kendall’s $\tau$ of {\bf 0.496} —an improvement of over 190.06\% compared to its performance on the raw corpus (\textbf{0.171}).

\begin{table}[ht!]
\centering
\begin{tabular}{|c|c|c|c|c|c|}
\hline
  \multirow{2}{*}{$D$} & \multicolumn{2}{c|}{\bf Word2Vec} & \multicolumn{2}{c|}{\bf FastText} & \multirow{2}{*}{\bf Glove}\\ 
  \cline{2-5}
  &CBOW & Skip-Gram & CBOW  & Skip-Gram & \\ 
  \cline{1-6}
  \hline  
  300 & 0.243 & 0.295 &0.470 & 0.408 & \bf  \color{blue} 0.496\\ \hline  
 \end{tabular}
 \caption{Word Semantic Similarity, Kendall’s $\tau$: Static LM trained on $\pi$-{\sc yalli} {\bf enhanced preprocessing}.} 
 \label{tab:tau_test_UNI+}
\end{table}

\subsection{Task II: Semantic Similarity Between Sentences}
We introduce here the task of semantic similarity between sentences.
Indeed, when computing semantic similarity between words, large language models (LLMs) may have access\footnote{It is not possible to guarantee that LLMs do not access external resources on the Internet, even if explicitly prohibited in the prompt. In our setup, we allowed the LLMs to decide autonomously whether or not to use such external access.} to privileged resources, such as dictionaries in various languages.

For example, when estimating the semantic similarity between a pair of words $(X, Y)_{NA}$ in Nawatl, which may be unknown to an LLM, it is likely that the definition $d$ of $X$ could be in a Nahuatl-French dictionary ($d_{\rm NA:FR}(X) \rightarrow$ French definition) and likewise the definition $d_{\rm NA:EN}(Y)$ should be found in a Nawatl-English dictionary. The dictionaries may have been absorbed during the LLM's vast learning process. In this way, the LLM is limited to comparing the representation of the definitions in French  $d_{\rm NA:FR}(X)$ and in English $d_{\rm NA:EN}(Y)$.
As a result, the LLM could bypass the native Nawatl words $(X, Y)_{\mathrm{NA}}$ altogether and simply compare the definitions in French and English,\footnote{Perhaps by simply computing the cosine similarity between the embeddings of the two definitions: $sim=\cos( \vec{E}(d_{\rm NA:FR}) \vec{E}(d_{\rm NA:EN}))$.} without directly using the words $(X,Y)_{NA}$ in Nawatl.
To avoid this situation, the new semantic evaluation protocol we propose is now performed between pairs of sentences as opposed to the previous task where the comparison was performed between word pairs.
There are few complete sentences to be found in dictionaries.
A total of $30$ sentences were written in Spanish by an annotator. 
These $30$ sentences were divided into 6 blocks of 5 sentences each. Each block was assigned to an annotator who drafted 5 candidate sentences with varying semantic similarity to the original sentence. 
All sentences (references and candidates) were translated into the Central Nawatl variant by a bilingual Nahuaphone (Spanish-Nawatl).

\begin{table}[ht!]
\centering
 \begin{tabular}{|r|c|c|c|c|}
  \hline
  & \bf Sentences & \bf Tokens & \bf Types &\bf Mean  \\ \hline
  \bf References& 30 &246 & 189 & 8.10 \\ \hline
  \bf Candidates& 150 &1026 & 599  & 6.84 \\ \hline
 \end{tabular}
\caption{The basic statistics for the task II.}
\end{table}

Appendix A shows one example of reference–candidate sentence blocks that are included in this reference ranking ($R_R$).
The reference ranking $R_R$ is produced by humans, and the goal of the task is to measure how closely the rankings $R_M$ generated by various language models approximate the reference ranking $R_R$. The rank correlation coefficient is again Kendall's $\tau(R_R, R_M)$.
In this experiment, we used the following LLMs via their API\footnote{
The API accesses were performed on 09/06/2025 from 16h-20h GMT. Interactive accesses (ChatGPT, Copilot, Grok and Claude) were performed on June 9-10/2025.}: Gemini-2.5-flash-preview-05-20; DeepSeek-V3-0324; Llama-3.1-70B-Instruct and Mistral-lar\-ge-latest. In interactive chat mode we have used ChatGPT, Copilot Grok and Claude.
%

%
The prompt was written in French, references and candidates in Central Nawatl: 
<<{ \it
\noindent \`A partir de la phrase nahuatl: ``[référence.]'' triez par ordre sémantique, de la plus proche à la plus lointaine, les cinq phrases suivantes: ``[candidate$_1$.]''; ``[candidate$_2$.]''; ``[candi\-date$_3$.]''; ``[candidate$_4$.]''; ``[candi\-date$_5$.]''.  Ne donnez pas des rankings de phrases autres que celles proposées. }>>
Table \ref{tab:tau_phrases_LLM} shows the results of LLMs, FastText and LSA on the semantic task of inter-sentence similarity. 
The best model is Gemini 2.5-flash-preview-05-20, with a Ken\-dall's $\tau=0.653$, but FastText and LSA are still competitive in this task.

\begin{table}[ht!]
  \centering
  \begin{tabular}{|c|c|}
\hline
  {\bf LLMs, FastText ($\pi$-yalli) and LSA} & \bf ~Kendall's $\tau$~\\ \hline
  Gemini 2.5-flash-preview-05-20 &\bf 0.653  \\ \hline  
   Claude 3.7 & 0.566 \\ \hline
  DeepSeek V3-0324 &  0.500 \\ \hline
\it \textbf{FastText 300D unigraphy + 4 stopwords}&\it \textbf{0.487} \\ \hline
  Grok 3& 0.486  \\ \hline
  Copilot & 0.446  \\ \hline
\it  \textbf{LSA 136D WLSE unigraphy +  4 stopwords} & \it \textbf{0.400} \\ \hline
  ChatGPT-4 mini & 0.386 \\  \hline 
\bf \textit{FastText 300D raw text}&\it  \bf \textit{0.373} \\ \hline
\bf \textit{LSA 165D WLSE unigraphy} & \bf \textit{0.367}\\ \hline
\bf \textit{LSA 139D WLSE raw text}& \bf \textit{0.360} \\ \hline
  Mistral Large-latest& 0.360 \\ \hline
  Llama-3.1-70B-Instruct& 0.219  \\ \hline
  \end{tabular}  
   \caption{Sentence Semantic Similarity Task, Kendall’s $\tau$: LLMs vs FastText and LSA. Stopwords eliminated: \textit{iwan (and), in (the), tlen (what)}, and \textit{ipan (in, over)}.} 
 \label{tab:tau_phrases_LLM}
\end{table}

\begin{figure}[hbt!]
    \centering
    \begin{tikzpicture}
        \begin{axis}[
            width=11.5cm,
            height=5.5cm,
            ylabel={Average Kendall's $\tau$ value},
            ylabel style={yshift=0.0002cm, xshift=10pt},
            ymin=.05,
            ymax=0.4,
            xticklabel style={rotate=45, anchor=east},
            ymajorgrids=true,
            grid style=dashed,
            xtick=data,
            xticklabel style={font=\scriptsize, rotate=45, anchor=east},
            symbolic x coords={1,2,10,20,30,40,50,70,85,87,88,90,94,95,96,97,98,99,100,101,104,116,123,139,154,167,200,300,400,500,600,602,610},
            enlarge x limits=0.03,
            legend entries={Embedding's addition, Lineal combination embeddings},
            legend style={at={(1,0)}, anchor=south east, legend columns=1} 
        ]
        \addplot[
            color=blue,
            line width=1.5pt
        ]
        coordinates {
            (2, 0.127)
            (10, 0.160)
            (20, 0.187)
            (30, 0.227)
            (40, 0.240)
            (50, 0.247)
            (70, 0.233)
            (85, 0.240)
            (87, 0.300)
            (88, 0.293)
            (90, 0.300)
            (94, 0.280)
            (95, 0.253)
            (96, 0.320)
            (97, 0.327)
            (98, 0.307)
            (99, 0.253)
            (100, 0.287)
            (101, 0.313)
            (104, 0.313)
            (116, 0.293)
            (123, 0.300)
            (139, 0.240)
            (154, 0.227)
            (167, 0.233)
            (200, 0.240)
            (300, 0.240)
            (400, 0.240)
            (500, 0.240)
            (600, 0.240)
            (602, 0.240)
        };
        \addplot[
            color=orange,
            line width=1.5pt,
        ]
        coordinates {
            (2, 0.100)
            (10, 0.200)
            (20, 0.167)
            (30, 0.193)
            (40, 0.220)
            (50, 0.200)
            (70, 0.220)
            (85, 0.187)
            (87, 0.287)
            (88, 0.247)
            (90, 0.233)
            (94, 0.220)
            (95, 0.220)
            (96, 0.273)
            (97, 0.253)
            (98, 0.280)
            (99, 0.247)
            (100, 0.260)
            (101, 0.293)
            (104, 0.287)
            (116, 0.333)
            (123, 0.327)
            (139, 0.360)
            (154, 0.353)
            (167, 0.353)
            (200, 0.287)
            (300, 0.287)
            (400, 0.287)
            (500, 0.287)
            (600, 0.287)
            (602, 0.287)
        };
        \end{axis}
    \end{tikzpicture}
%
    \begin{tikzpicture}
        \begin{axis}[
            width=11.5cm,
            height=5.5cm,
            xlabel={Number of LSA components},
            ylabel={Average Kendall's $\tau$ value},
            ylabel style={yshift=0.0002cm, xshift=10pt},
            ymin=.05,
            ymax=0.4,
            xticklabel style={rotate=45, anchor=east},
            ymajorgrids=true,
            grid style=dashed,
            xtick=data,
            symbolic x coords={2,10,20,30,40,50,60,70,80,87,88,89,90,96,98,100,104,115,126,141,146,150,154,156,163,164,165,167,168,200,300,400,500,596},
            xticklabel style={font=\scriptsize, rotate=45, anchor=east},
            enlarge x limits=0.03,
            legend entries={Embedding's addition, Lineal combination embeddings},
            legend style={at={(1,0)}, anchor=south east, legend columns=1} 
        ]
        \addplot[
            color=blue,
            line width=1.5pt
        ]
        coordinates {
            (2, 0.080)
            (10, 0.160)
            (20, 0.213)
            (30, 0.187)
            (40, 0.280)
            (50, 0.187)
            (60, 0.227)
            (70, 0.240)
            (80, 0.253)
            (87, 0.327)
            (88, 0.280)
            (89, 0.287)
            (90, 0.307)
            (96, 0.280)
            (98, 0.307)
            (100, 0.300)
            (104, 0.287)
            (115, 0.313)
            (126, 0.267)
            (141, 0.247)
            (146, 0.220)
            (150, 0.253)
            (154, 0.247)
            (156, 0.253)
            (163, 0.240)
            (164, 0.247)
            (165, 0.240)
            (167, 0.233)
            (168, 0.240)
            (200, 0.233)
            (300, 0.233)
            (400, 0.233)
            (500, 0.233)
            (596, 0.233)
        };
        \addplot[
            color=orange,
            line width=1.5pt,
        ]
        coordinates {
            (2, 0.093)
            (10, 0.173)
            (20, 0.207)
            (30, 0.180)
            (40, 0.233)
            (50, 0.193)
            (60, 0.207)
            (70, 0.220)
            (80, 0.240)
            (87, 0.320)
            (88, 0.240)
            (89, 0.233)
            (90, 0.273)
            (96, 0.260)
            (98, 0.313)
            (100, 0.293)
            (104, 0.320)
            (115, 0.313)
            (126, 0.353)
            (141, 0.353)
            (146, 0.347)
            (150, 0.333)
            (154, 0.340)
            (156, 0.347)
            (163, 0.347)
            (164, 0.347)
            (165, 0.367)
            (167, 0.347)
            (168, 0.340)
            (200, 0.300)
            (300, 0.300)
            (400, 0.300)
            (500, 0.300)
            (596, 0.300)
        };
        \end{axis}
    \end{tikzpicture}
    \caption{Relationship between number of LSA components (in blue embeddings addition, in orange lineal addition of embeddings) and average Kendall's $\tau$ (raw text sentences (top) and with unigraphy (bottom)).}
    \label{fig:lsacomcor}
\end{figure}
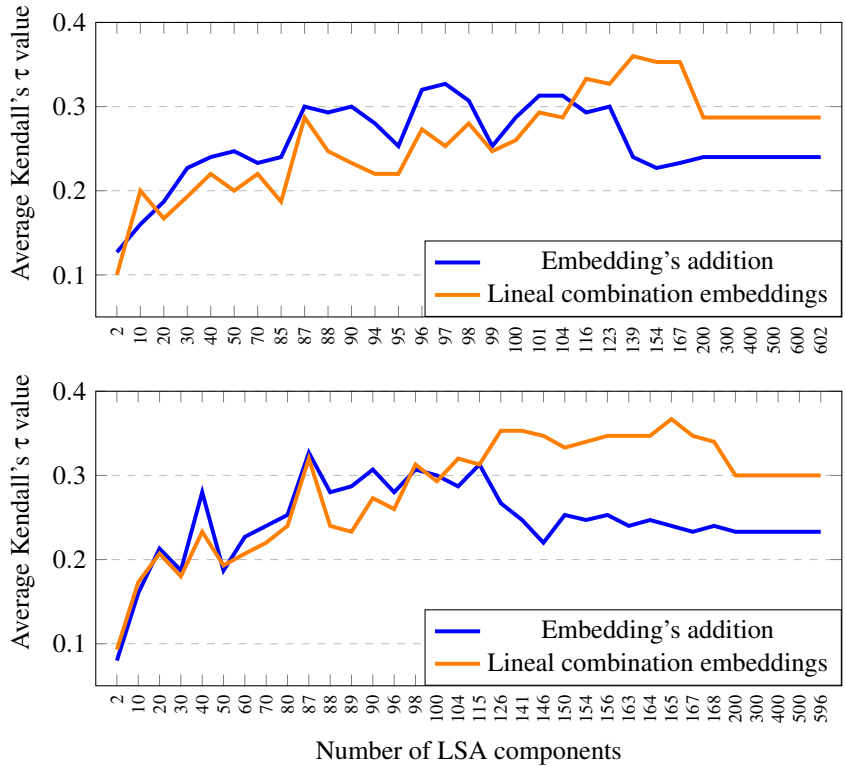

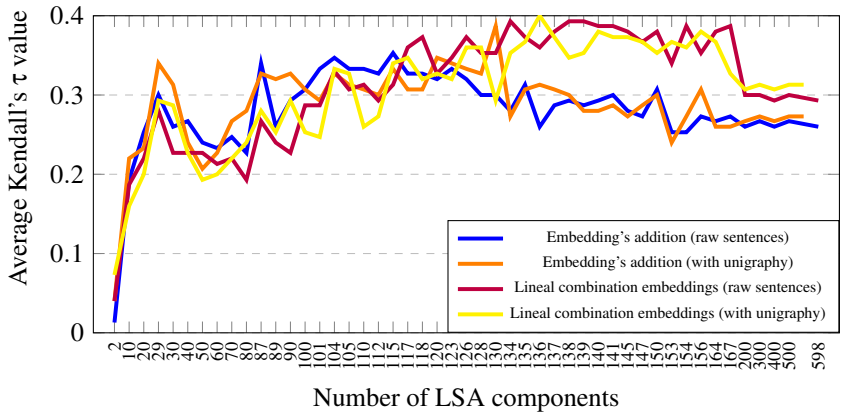
\begin{figure}[hbt!]
    \centering
    \begin{tikzpicture}
        \begin{axis}[
            width=11.5cm,
            height=5.8cm,
            xlabel={Number of LSA components},
            ylabel={Average Kendall's $\tau$ value},
            ylabel style={yshift=0.0002cm, xshift=10pt},
            ymin=-.0,
            ymax=0.4,
            xticklabel style={rotate=45, anchor=east},
            ymajorgrids=true,
            grid style=dashed,
            xtick=data,
            symbolic x coords={2,10,20,29,30,40,50,60,70,80,87,89,90,100,101,104,105,110,112,115,117,118,120,123,126,128,130,134,135,136,137,138,139,140,141,145,147,150,153,154,156,164,167,200,300,400,500,592,598},
            xticklabel style={font=\scriptsize, rotate=45, anchor=east},
            enlarge x limits=0.03,
            legend entries={Embedding's addition (raw sentences), Embedding's addition (with unigraphy), Lineal combination embeddings (raw sentences), Lineal combination embeddings (with unigraphy)},
            legend style={at={(1,0)}, anchor=south east, legend columns=1, font=\tiny} 
        ]
        \addplot[
            color=blue,
            line width=1.5pt
        ]
        coordinates {
            (2, 0.013)
            (10, 0.193)
            (20, 0.253)
            (29, 0.300)
            (30, 0.260)
            (40, 0.267)
            (50, 0.240)
            (60, 0.233)
            (70, 0.247)
            (80, 0.227)
            (87, 0.340)
            (89, 0.260)
            (90, 0.293)
            (100, 0.307)
            (101, 0.333)
            (104, 0.347)
            (105, 0.333)
            (110, 0.333)
            (112, 0.327)
            (115, 0.353)
            (117, 0.327)
            (118, 0.327)
            (120, 0.320)
            (123, 0.333)
            (126, 0.320)
            (128, 0.300)
            (130, 0.300)
            (134, 0.280)
            (135, 0.313)
            (136, 0.260)
            (137, 0.287)
            (138, 0.293)
            (139, 0.287)
            (140, 0.293)
            (141, 0.300)
            (145, 0.280)
            (147, 0.273)
            (150, 0.307)
            (153, 0.253)
            (154, 0.253)
            (156, 0.273)
            (164, 0.267)
            (167, 0.273)
            (200, 0.260)
            (300, 0.267)
            (400, 0.260)
            (500, 0.267)
            (598, 0.260)
        };
        \addplot[
            color=orange,
            line width=1.5pt,
        ]
        coordinates {
            (2, 0.040)
            (10, 0.220)
            (20, 0.233)
            (29, 0.340)
            (30, 0.313)
            (40, 0.240)
            (50, 0.207)
            (60, 0.227)
            (70, 0.267)
            (80, 0.280)
            (87, 0.327)
            (89, 0.320)
            (90, 0.327)
            (100, 0.307)
            (101, 0.293)
            (104, 0.327)
            (105, 0.313)
            (110, 0.307)
            (112, 0.300)
            (115, 0.333)
            (117, 0.307)
            (118, 0.307)
            (120, 0.347)
            (123, 0.340)
            (126, 0.333)
            (128, 0.327)
            (130, 0.387)
            (134, 0.273)
            (135, 0.307)
            (136, 0.313)
            (137, 0.307)
            (138, 0.300)
            (139, 0.280)
            (140, 0.280)
            (141, 0.287)
            (145, 0.273)
            (147, 0.287)
            (150, 0.300)
            (153, 0.240)
            (154, 0.273)
            (156, 0.307)
            (164, 0.260)
            (167, 0.260)
            (200, 0.267)
            (300, 0.273)
            (400, 0.267)
            (500, 0.273)
            (592, 0.273)
        };
        \addplot[
            color=purple,
            line width=1.5pt,
        ]
        coordinates {
            (2, 0.040)
            (10, 0.187)
            (20, 0.220)
            (29, 0.280)
            (30, 0.227)
            (40, 0.227)
            (50, 0.227)
            (60, 0.213)
            (70, 0.220)
            (80, 0.193)
            (87, 0.267)
            (89, 0.240)
            (90, 0.227)
            (100, 0.287)
            (101, 0.287)
            (104, 0.33)
            (105, 0.307)
            (110, 0.313)
            (112, 0.293)
            (115, 0.313)
            (117, 0.360)
            (118, 0.373)
            (120, 0.327)
            (123, 0.347)
            (126, 0.373)
            (128, 0.353)
            (130, 0.353)
            (134, 0.393)
            (135, 0.373)
            (136, 0.360)
            (137, 0.380)
            (138, 0.393)
            (139, 0.393)
            (140, 0.387)
            (141, 0.387)
            (145, 0.380)
            (147, 0.367)
            (150, 0.380)
            (153, 0.340)
            (154, 0.387)
            (156, 0.353)
            (164, 0.380)
            (167, 0.387)
            (200, 0.300)
            (300, 0.300)
            (400, 0.293)
            (500, 0.300)
            (598, 0.293)
        };
        \addplot[
            color=yellow,
            line width=1.5pt,
        ]
        coordinates {
            (2, 0.073)
            (10, 0.160)
            (20, 0.200)
            (29, 0.293)
            (30, 0.287)
            (40, 0.227)
            (50, 0.193)
            (60, 0.200)
            (70, 0.220)
            (80, 0.240)
            (87, 0.280)
            (89, 0.253)
            (90, 0.293)
            (100, 0.253)
            (101, 0.247)
            (104, 0.333)
            (105, 0.327)
            (110, 0.260)
            (112, 0.273)
            (115, 0.340)
            (117, 0.347)
            (118, 0.320)
            (120, 0.327)
            (123, 0.320)
            (126, 0.360)
            (128, 0.360)
            (130, 0.293)
            (134, 0.353)
            (135, 0.367)
            (136, 0.400)
            (137, 0.373)
            (138, 0.347)
            (139, 0.353)
            (140, 0.380)
            (141, 0.373)
            (145, 0.373)
            (147, 0.367)
            (150, 0.353)
            (153, 0.367)
            (154, 0.360)
            (156, 0.380)
            (164, 0.367)
            (167, 0.327)
            (200, 0.307)
            (300, 0.313)
            (400, 0.307)
            (500, 0.313)
            (592, 0.313)
        };
        \end{axis}
    \end{tikzpicture}
    \caption{Relationship between number of LSA components (embeddings addition, lineal addition of embeddings, raw text sentences and with unigraphy) and average Kendall's $\tau$  without 4 stopwords ('\textit{iwan}', '\textit{in}', '\textit{tlen}', '\textit{ipan}').}
    \label{fig:lsa4STOP}
\end{figure}

The static models used were as follows: Word2Vec \cite{Mikolov2013distributed}, FastText~ \cite{bojanowski-etal-2017-enriching}, and Glove~ \cite{glove}. 
\color{black}
We also tested the classical LSA method \cite{lsa} which does not
use learning but matrix decomposition by SVD.
\color{black}
Table \ref{tab:tache2} shows the results of these models on the sentence semantic similarity task.
In this table, the raw text column shows the results 
using from the corpus without orthographic unification measures (unigraphy).
The best model is Word2Vec Skip-Gram which obtains a $\tau$=0.386.
Finally, the Unigraphied column of the table \ref{tab:tache2} shows the results of the same models but trained on the $\pi$-\textsc{yalli} corpus with the enhanced unified orthography.
The best model is FastText Skip-Gram with $\tau$=\textbf{0.487}, an increase of approximately 30.56\% over the same model trained on raw text.
The Fig. \ref{fig:lsacomcor} and Fig. \ref{fig:lsa4STOP} shows the results of applying LSA using a restricted vocabulary composed only of terms from the semantic task ($\approx$ 600 terms). In both cases, experiments were conducted with and without the application of unigraphy.
The Fig. \ref{fig:lsacomcor} represent the results obtained without removing any stopwords. The best performance ($\tau$ = \textbf{0.367}), surpassing that of Llama and Mistral, was achieved using 165 latent dimensions, applying unigraphy, and representing sentences as a weighted linear sum of embeddings (WLSE).
The Fig. \ref{fig:lsa4STOP} extends this analysis by additionally removing 4 stopwords identified in Nawatl: \textit{iwan (and), in (the), tlen (what),} and \textit{ipan (in, over)}. Under these conditions, the highest score ($\tau$ = \textbf{0.400}), which outperforms Llama, Mistral, and Chat-GPT, was obtained using 136 latent dimensions, also employing unigraphy and WLSE for sentence construction. The positional embeddings \cite{wang-chen-2020-position} do not seem to work on this task.

\begin{table}[ht!]
    \centering
    \begin{tabular}{|c|c|c|}
        \multicolumn{3}{c}{\bf Local LM trained on $\pi$-\textsc{yalli} corpus} \\ 
      \hline
        \bf Algorithm &\bf ~Raw text &\bf Unigraphied \\ \hline        
        Word2Vec CBOW & 0.300 & 0.353 \\\hline
        Word2Vec Skip-Gram & 0.293 & 0.347 \\ \hline
        FastText CBOW & 0.327 & 0.347 \\ \hline
        \bf FastText Skip-Gram & 0.373 & \bf 0.487 \\ \hline
        Glove  & 0.160 & 0.207 \\ \hline
    \end{tabular}
    \caption{Sentence Semantic Similarity, Kendall’s $\tau$: Raw vs. Unigraphied corpus.}
    \label{tab:tache2}
\end{table}

\section{Discussion and conclusions}
\label{sec:CONC}

The main objective of this article is to show that unigraphy makes it possible to obtain a standardized version of the Nawatl text to be better processed by machines. This standardization helps models trained on a unigraphied corpus achieve better performance in NLP tasks.
In particular, we have confirmed this in the computation of semantic similarity between words and between sentences.
We also found that CBOW versions (in both Word2Vec and FastText models) perform better for Task I (word similarity), while the Skip-Gram versions perform better for Task II (sentence similarity). 
This can be explained as follows:
the CBOW architecture is designed to predict a word from a set of context words, whereas Skip-Gram predicts a set of context words from a single target word. CBOW performs better in Task I because that task is based on word similarity. On the other hand (using only a vocabulary of 139 types), Skip-Gram was better for Task II due to its focus on context prediction.
%
%
The FastText Skip-Gram model trained on $\pi$-{\sc yalli} outperforms  Llama-3.1-70B-Instruct, Mistral Large-latest, ChatGPT-4 mini and Copilot. It is at a comparable level to Grok 3, but is less good than  DeepSeek V3, Claude and Gemini-2.5. On the other hand, LSA outperforms Llama-3.1 and Mistral Large.
Our results showed that static Nawatl models and classic statistical ones using unified orthography, remain competitive compared to the most powerful large language models.
This is especially true when processing languages like Nawatl, which have very limited computational resources.
For future work, given the results, it is planned to continue adding and modifying the rules in the symbolic model for unification of Nawatl. As well as adding statistical or neural models and other measures to compare their performances. The code is available in a GitHub repository for anyone interested in testing their Nawatl texts \footnote{Code available: \url{https://demo-lia.univ-avignon.fr/pi-yalli/}}.



\section*{Appendix A. Semantic similarity of sentences}

We present one example of the sentences used for the similarity calculation. The candidate sentences are ordered from highest to lowest semantic similarity with respect to the reference.\\ 

\noindent 
~\\
\textsc{ Reference$_1$:} \textrm{\color{red}{Inn xochitlahkuilolli powi itech intetlasohtlalis ome ichpochtli, telpochtli tenamikwan}} / \textit{{\color{Bittersweet}The novel is based on the love of a young couple}}

\noindent \textsc{Candidates:}
\color{blue}
    \begin{enumerate}
    \item Inin sasanilli kipowa intetlasohtlalis ichpochtli iwan telpochtli. / \color{black}\color{Bittersweet}\textit{This story is about the love of a girl and a boy}\color{blue}
    \item Tetlasohtalistli keman se okatka ichpochtli, telpochtli miakpa mopowa ipan xochitlahkuilolli. /\color{Bittersweet}\textit{Young love is often told in literature}
\color{blue} \item Ihkuak se ichpochtli, se tepochtli, tetlasohtlalistli kualtzin. /
 \color{Bittersweet}\textit{The youth love is beautiful}
\color{blue}
    \item Tetlasohtlalistli iwan telpochkayotl, ichpochkayotl nochipa ome tlahtolli sepanitokeh. /
\color{Bittersweet} \textit{Love and youth are a couple that go hand in hand}\color{blue}
    \item Ok achi kualli se miki ihkuak ichpochtli, telpochtli, wan ihkin amo se tetlasohtla. / 
\color{Bittersweet} \textit{Better to die young than to fall in love}
\end{enumerate}~

\color{black}

\section{Credits}
This study was funded by Avignon Université, France (grant Agorantic/Intermedius Projects NAHU and NAHU²).

\bibliographystyle{apalike}
\bibliography{references.bib}

\begin{thebibliography}{}

\bibitem[Abdillahi et~al., 2006]{abdillahi:hal-01311495}
Abdillahi, N., Nocera, P., and Torres, J.~M. (2006).
\newblock {Boites a outils TAL pour les langues peu informatis{\'e}es : Le cas
  du Somali}.
\newblock In {\em JADT}, Besan{\c c}on, France.

\bibitem[Aguilar~Santiago and Garc{\'\i}a~Z{\'u}{\~n}iga,
  2023]{aguilar2023tecnologias}
Aguilar~Santiago, C.~A. and Garc{\'\i}a~Z{\'u}{\~n}iga, H.~A. (2023).
\newblock Tecnolog\'ias del lenguaje aplicadas al procesamiento de lenguas
  ind\'igenas en {M}éxico: Una visi{\'o}n general.
\newblock {\em Ling{\"u}{\'\i}stica y Literatura}, (84):79--102.

\bibitem[Berment, 2004]{these-pi}
Berment, V. (2004).
\newblock {\em Méthodes pour informatiser les langues et les groupes de
  langues ``peu dotées''}.
\newblock PhD thesis, Université Joseph-Fourier - Grenoble I.

\bibitem[Bojanowski et~al., 2017]{bojanowski-etal-2017-enriching}
Bojanowski, P., Grave, E., Joulin, A., and Mikolov, T. (2017).
\newblock Enriching word vectors with subword information.
\newblock {\em Transactions of the ACL}, 5:135--146.

\bibitem[de~Durand-Forest et~al., 1995]{parlons}
de~Durand-Forest, J., Dehouve, D., and Roulet, E. (1995).
\newblock {\em Parlons Nahuatl. La langue des Aztèques}.
\newblock L'Harmattan.

\bibitem[Deerwester et~al., 1990]{lsa}
Deerwester, S., Dumais, S.~T., Furnas, G.~W., Landauer, T.~K., and Harshman, R.
  (1990).
\newblock Indexing by latent semantic analysis.
\newblock {\em Journal of the American Society for Information Science},
  41(6):391--407.

\bibitem[Figueroa-Saavedra, 2024]{saavedra2024amapowalistli}
Figueroa-Saavedra, M. (2024).
\newblock {\em Amapowalistli iwan tlahkuilolewalistli. Tlamachtilamoxtli}.
\newblock Universidad Veracruzana.

\bibitem[Figueroa-Saavedra and Hern{\'a}ndez-Mart{\'\i}nez,
  2023]{saavedra2023nawatlahtolli}
Figueroa-Saavedra, M. and Hern{\'a}ndez-Mart{\'\i}nez, J.~{\'A}. (2023).
\newblock {\em In nawatlahtolli ipan interkoltoral tlamachtilistli itech
  Veracruz: owihkayotl iwan chikawakayotl}.
\newblock Universidad Veracruzana.

\bibitem[Guzm{\'a}n-Landa et~al., 2025]{piyalliTALN}
Guzm{\'a}n-Landa, J.-J., Torres-Moreno, J.-M., Ranger, G.,
  Garrido-Avenda{\~n}o, M.~L., Figueroa-Saavedra, M., Quintana-Torres, L.,
  Gonz{\'a}lez-Gallardo, C.-E., Linhares-Pontes, E., Vel{\'a}zquez-Morales, P.,
  and Moreno~Jim{\'e}nez, L.-G. (2025).
\newblock $\pi$-yalli: un nouveau corpus pour le nahuatl / \textit{Yankuik
  nawatlahtolkorpus pampa tlahtolmachiotl}.
\newblock In {\em TALN, Marseille}.

\bibitem[INALI, 2009]{INALI}
INALI (2009).
\newblock {\em Catálogo de las Lenguas Indígenas Nacionales}.
\newblock INALI, Mexico City.

\bibitem[INEGI, 2020]{inegi2020censo}
INEGI (2020).
\newblock {\em Censo de poblaci\'on y vivienda 2020}.
\newblock \url{https://www.
  inegi.org.mx/rnm/index.php/catalog/632/study-description}.

\bibitem[Mikolov et~al., 2013]{Mikolov2013distributed}
Mikolov, T., Sutskever, I., Chen, K., Corrado, G., and Dean, J. (2013).
\newblock Distributed representations of words and phrases and their
  compositionality.
\newblock In {\em NIPS-V2}, NIPS, page 3111–3119, Red Hook, NY.

\bibitem[Monzón, 1990]{monzon}
Monzón, C. (1990).
\newblock Registro de la variación fonológica en el náhuatl moderno. un
  estudio de caso.
\newblock In {\em CIESAS}. CIESAS.

\bibitem[{Moseley, Christopher (ed.)} and {Nicolas Alexandre (cart.)},
  2012]{unesco2012atlas}
{Moseley, Christopher (ed.)} and {Nicolas Alexandre (cart.)} (2012).
\newblock {\em Atlas des langues en danger dans le monde}.
\newblock UNESCO.

\bibitem[Pennington et~al., 2014]{glove}
Pennington, J., Socher, R., and Manning, C.~D. (2014).
\newblock Glove: Global vectors for word representation.
\newblock In {\em 2014 EMNLP}, pages 1532--1543. ACL.

\bibitem[Smith, 2002]{smith2002aztecs}
Smith, M. (2002).
\newblock {\em The Aztecs}.
\newblock Peoples of America. Wiley.

\bibitem[Torres-Moreno et~al., 2024]{NAHU2}
Torres-Moreno, J.-M., Avendaño-Garrido, M.-L., Figueroa-Saavedra, M., Ranger,
  G., González-Gallardo, C., Linhares~Pontes, E., Velazquez~Morales, P.,
  Quintana~Torres, L., and Guzm\'an-Landa, J.-J. (2024).
\newblock {NAHU}²: Un nouveau corpus pour le {N}ahuatl.
\newblock In {\em 18èmes J Informatique Centre-Val de Loire}.
  \url{https://hal.science/hal-04814636}.

\bibitem[Wang and Chen, 2020]{wang-chen-2020-position}
Wang, Y.-A. and Chen, Y.-N. (2020).
\newblock What do position embeddings learn? an empirical study of pre-trained
  language model positional encoding.
\newblock In {\em EMNLP}, pages 6840--6849. ACL.

\end{thebibliography}

\end{document}